# A Kriging-Random Forest Hybrid Model for Real-time Ground Property Prediction during Earth Pressure Balance Shield Tunneling

Ziheng Geng[1], Chao Zhang[2], Yuhao Ren[3], Minxiang Zhu[4], Renpeng Chen[5], and Hongzhan Cheng[6]

[1] Research assistant, College of Civil Engineering, Hunan University, Changsha 410082, China. Email: gengziheng@hnu.edu.cn
[2] Professor and corresponding author, Key Laboratory of Building Safety and Energy Efficiency of the Ministry of Education, Hunan University, Changsha 410082, China. Email: chao_zhang@hnu.edu.cn
[3] Research assistant, College of Civil Engineering, Hunan University, Changsha 410082, China. Email: renyuhao@hnu.edu.cn
[4] Research assistant, College of Civil Engineering, Hunan University, Changsha 410082, China. Email: zhuminxiang@hnu.edu.cn
[5] Professor, Research Center for Advanced Underground Space Technologies of Hunan University, Changsha 410082, China. Email: chenrp@hnu.edu.cn
[6] Associate professor, Research Center for Advanced Underground Space Technologies of Hunan University, Changsha 410082, China. Email: hzcheng@hnu.edu.cn

**Abstract**

A kriging-random forest hybrid model (KRF) is developed for real-time ground property prediction ahead of the earth pressure balanced (EPB) shield by integrating Kriging extrapolation and random forest, which can guide shield operating parameter selection thereby mitigate construction risks. The proposed KRF algorithm synergizes two types of information: prior information and real-time information. The previously predicted ground properties with EPB operating parameters are extrapolated via the Kriging algorithm to provide prior information for the prediction of currently being excavated ground properties. The real-time information refers to the real-time operating parameters of the EPB shield, which are input into random forest to provide a real-time prediction of ground properties. The integration of these two predictions is achieved by assigning weights to each prediction according to their uncertainties, ensuring the prediction of KRF with minimum uncertainty. The performance of the KRF algorithm is assessed via a case study of the Changsha Metro Line 4 project. It reveals that the proposed KRF algorithm can predict ground properties with an accuracy of 93%, overperforming the existing algorithms of LightGBM, AdaBoost-CART, and DNN by 29%, 8%, and 12%, respectively. Another dataset from Shenzhen Metro Line 13 project is utilized to further evaluate the model's generalization performance, revealing that the model can transfer its learned knowledge from one region to another with an accuracy of 89%.

# 1. Introduction

Earth pressure balance (EPB) shield has become a dominant tool for urban underground space construction over the past two decades, particularly in megacities like Shanghai, Tokyo, and Bangkok [1-3]. In these megacities, tunneling-induced ground disturbances, e.g., surface settlement, surface heave, and in-situ stress field alternation, can pose substantial risks to adjacent existing infrastructures, inducing damage even failure of lifeline infrastructures thereby hindering the resilience of megacities. Therefore, resilient city management calls for deliberate control of tunneling-induced ground disturbances. Physically, such disturbance is an outcome of the interaction between the ground and EPB shield, further dictated by the ground properties (e.g., rock types, cover depth, and groundwater condition), and operating parameters (e.g., thrust, torque, advance rate, cutter rotation speed, screw conveyor rotation speed, and foam generator). Therefore, the control of tunneling-induced ground disturbance requires a sophisticated strategy for adjusting the EPB operating parameters in response to changes in ground properties. Yet, to date, there is no established approach to directly monitor the properties of the ground currently being excavated, making it difficult to adjust the EPB operating parameters in real-time according to the current ground properties, thereby leading to a series of construction accidents like cutter wear and clogging [4,5], water or mud inflow [6,7], and tunnel excavation face collapse [8,9]. It is anticipated that these construction accidents can be largely mitigated if real-time predictions of ground properties are available.

Real-time prediction of ground properties from shield machine operating parameters via machine learning (ML) algorithm is recently emerging as an efficient method in tunneling engineering practice [10-13]. The underlying physics for these algorithms is that the machine operating parameters are intrinsic manifests of the machine–ground interaction and implicitly reflect the ground properties. Therefore, recent technical breakthroughs in artificial intelligence (AI) and ML can be harnessed for mining operating data to predict ground properties. The successful implementations include a support vector regression model proposed by Liu et al. [14]. It can effectively predict rock mass parameters, e.g., the uniaxial compressive strength (UCS), brittleness index (BI), etc., and their sudden changes. To improve the computational efficiency of the support vector machine, Zhang et al. [15] utilized a dimensionality reduction algorithm to compress the TBM operating data and then employed a support vector classifier to predict the rock mass type. With the knowledge that integrating multiple learners can generally yield better performance than a single learner, Liu et al. [16] presented an ensemble learning model underpinned by classification and regression tree and AdaBoost algorithm to classify the surrounding rock mass. In addition, Zhao et al. [17] and Yu and Mooney [18] developed a multiple-output artificial neural network and a semi-supervised learning model, respectively, to predict the multiple geological types contained in the tunnel cross-section. However, these applications are predominantly confined to hard rock tunnel boring machines (TBM). Real-time ground property prediction of EPB shield remains largely unexplored, despite its wide adoption in urban space development.

Herein, an algorithm called the kriging-random forest hybrid model (KRF) is proposed for the real-time prediction of ground properties during EPB tunneling. The main novelty of this algorithm lies in its ability to utilize not only real-time operating parameters but also prior information. The previously predicted ground properties are extrapolated via the Kriging algorithm to provide a prior estimation for the real-time ground property prediction. The EPB operating parameters are utilized as input to the random forest to provide a real-time prediction of ground properties. These two predictions are integrated via the weighted least squares method, which assigns weights inversely proportional to their relative uncertainties. The applicability of the proposed algorithm is assessed with a dataset collected from a metro project in Changsha. It demonstrates that the KRF algorithm, incorporating both prior and real-time information, can predict ground properties with higher accuracy than existing data-driven models. The generalization performance of the proposed algorithm is further evaluated on another dataset collected from Shenzhen city, demonstrating its strong adaptability to geology in different regions.

## 2. Prior and Real-time Information

The EPB tunneling process involves two types of information: prior and real-time information, shown in Fig. 1. Prior information is provided by the Kriging extrapolations of the previously predicted ground properties. The parameters in Kriging extrapolation involve range, nugget, and sill, which can be calibrated by the regional geological analysis. The previously excavated ground properties can be predicted from either EPB operating parameters or muck analysis. The significance of such information to current ground properties prediction is embedded in the spatial correlation among ground properties within a given region. Real-time information is provided by the current EPB operating parameters. This information implicitly reflects the real-time characteristics of the shield–ground interaction, thereby reflecting the mechanical properties of the currently being excavated ground. How the two types of information relate to current ground properties is elaborated below.

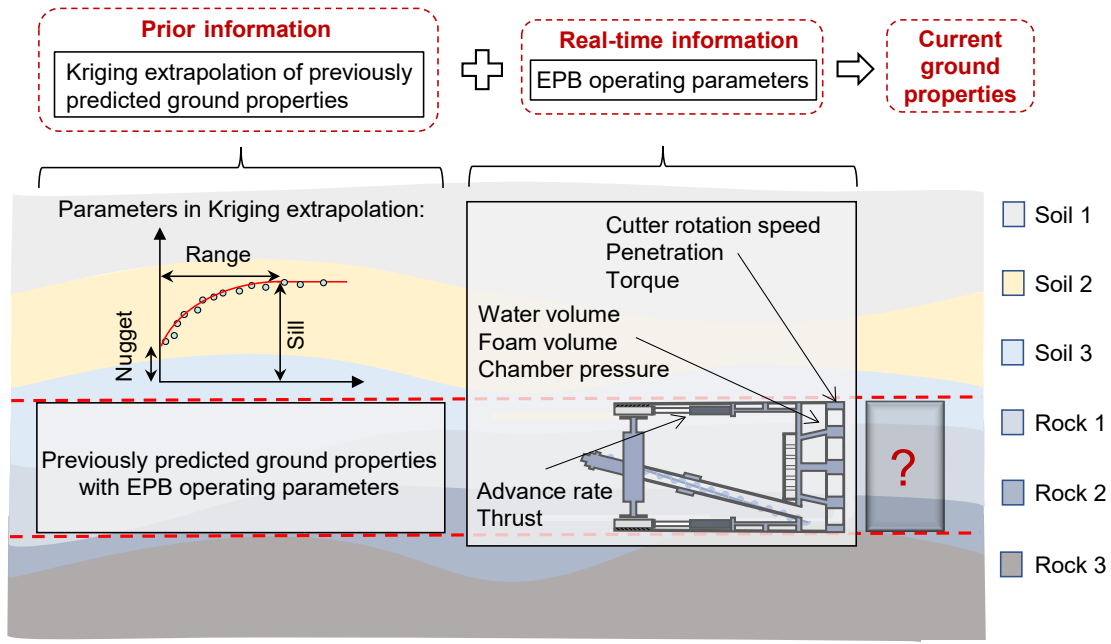

**Fig. 1.** Integration of prior and real-time information for predicting ground properties ahead of the EPB shield.

Prior information incorporates the inherent spatial correlation of ground properties among the current location and previously excavated locations. Due to the continuity of geological processes, ground properties tend to exhibit spatial correlation, manifesting as a stronger dependence for two closer points [19]. Thus, the ground type and properties vary with the tunnel axis in a certain continuous manner if there is no fault encountered, illustrated in Fig. 1. This continuous manner is encoded by the geological history in the region of interest, referred to as the geostatistical characteristic. This inherent geostatistical characteristic suggests that the ground properties along the tunnel axis exhibit predefined spatial correlation and conform to a particular geostatistical distribution function. In practice, this spatial correlation is represented by a distance function. Note that this geostatistical distribution function is an inherent characteristic at the regional scale. That is, the geostatistical distribution function is prescribed by the geological history and can be calibrated by regional geological analysis, thus can serve as prior information for ground property prediction. This function facilitates the prediction of currently being excavated ground properties by extrapolating the properties of the previously excavated ground, providing a prior estimation for real-time prediction of ground properties.

Real-time information refers to the real-time operating parameters of the EPB shield. The magnitude and variation of these parameters are manifests of real-time shield–ground interactions in multiple aspects. They can be classified into three categories according to the underlying mechanical processes: hydraulic cylinder, cutterhead, and chamber. The hydraulic cylinder parameters include thrust and advance rate, quantifying the axial resistances provided by the ground thereby tacitly correlating to the mechanical properties of the currently being excavated ground. The cutterhead parameters, including cutter rotation speed, penetration, and torque, represent the efficiency in breaking soil and rocks, reflecting the hardness and strength of the currently being excavated ground. The chamber parameters consist of pressure, and the volume of foam and water, depending on the ground properties of mineralogy and size distribution. Therefore, it is feasible to mine data on these operating parameters and provide a real-time prediction of ground properties.

As elaborated above, both prior and real-time information are correlated with ground properties, thereby can be utilized for ground property prediction. Yet, only real-time information is included in existing data-driven models of ground property prediction for tunnel excavation via EPB or other TBMs. The reliability of these data-driven models highly relies on the validity of the data, e.g., quantity, quality, and diversity. To date, an inclusive dataset is still lacking, partly due to the significant variability of geology worldwide. Consequently, the model based on limited data inexorably contains bias, which can be potentially mitigated by improving the data validity. Prior information originates from regional geological analysis and is therefore expected to supplement real-time information to some extent. Its integration into a dataset can better constrain the predictions of data-driven models from the viewpoint of geostatistics. Under this premise, a KRF algorithm was developed to harness both prior and real-time information for ground property prediction.

## 3. Development of Kriging-Random Forest Hybrid model

The development of the KRF algorithm consists of three procedures, shown in Fig. 2. In the first procedure, the Kriging algorithm is utilized to extrapolate the current ground properties from the previously excavated ground properties. In the second procedure, the random forest algorithm is employed to predict the current ground properties from real-time EPB operating parameters. The third procedure assigns weights to the Kriging extrapolation and random forest prediction according to their relative uncertainties, integrating the two predictions via a weighted least squares (WLS) method. It is expected that this integration can constrain random forest predictions with the aid of the spatial correlation of ground properties, enhancing the accuracy of real-time ground property prediction. The workflow of the three procedures is elaborated below.

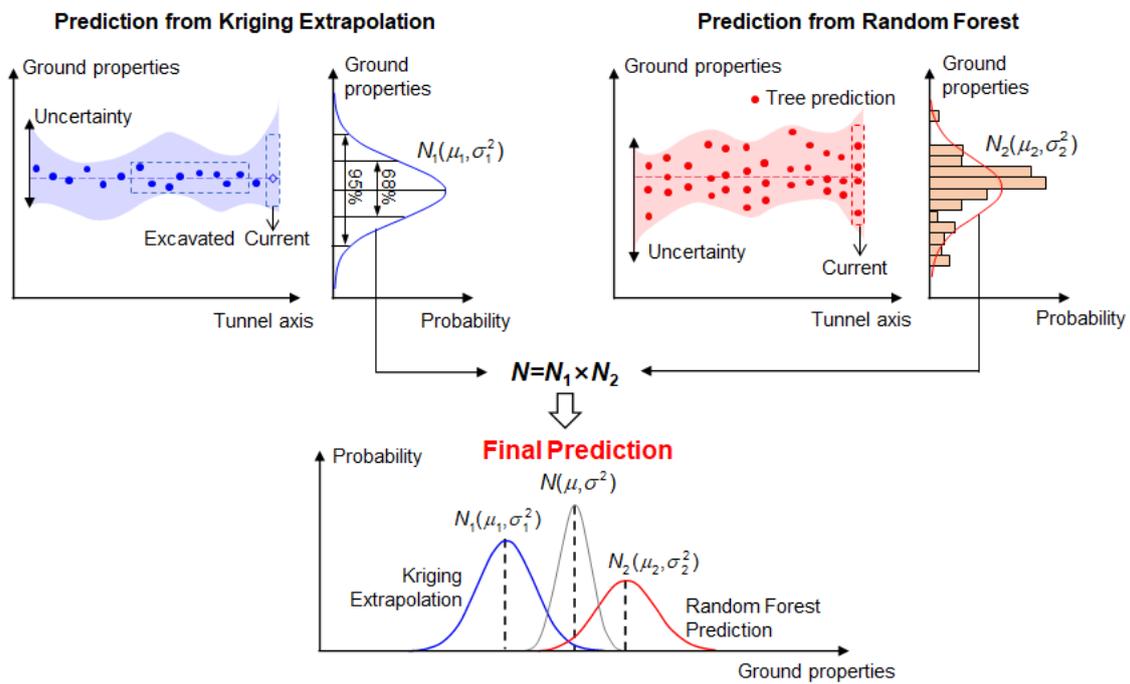

**Fig. 2.** Development of kriging-random forest hybrid model.

## 3.1. Adopting Kriging extrapolation to estimate current ground properties

Several spatial extrapolation techniques are available to establish a continuous distribution of ground properties with discrete points, e.g., inverse distance weighting [20], spline [21], and Kriging [19]. Among these techniques, the inverse distance weighting and spline techniques rely only on the geometric properties of known ground properties. By contrast, the Kriging algorithm incorporates both the geometric properties and spatial correlation of known ground properties. The inclusion of spatial correlation allows the Kriging algorithm to produce a more reliable estimation, particularly when data sampling is sparse. Therefore, the Kriging algorithm has been widely adopted in practice [22-24]. Here, the Kriging algorithm was selected to extrapolate the current ground properties from previously excavated ground properties. This is implemented in three steps: exploratory data analysis (EDA), semi-variogram modeling, and spatial extrapolation.

In the first step, EDA eliminates outliers in the raw data from regional geological analysis and converts the rest of the data into a normally distributed dataset. Generally, the raw data from regional geological analysis contains several data points that abnormally deviate from other data values, resulting in model misspecification and biased estimation [25]. Here, a standard procedure suggested by Cattle et al. [26] was selected to remove data points beyond three standard deviations from the mean values. Subsequently, the Kolmogorov–Smirnov test was implemented to check whether the data distribution was normal; otherwise, a logarithmic transformation was performed to convert the dataset to be normally distributed [23,27].

The second step calibrates the semi-variogram model, which quantifies the spatial variation in ground properties as a distance function. The semi-variogram model is a function typically fitted by the variation data points. These data points of variation $\gamma^*(h)$ can be calculated from the dataset from the first step, expressed as:

$$\gamma^*(h) = \frac{1}{2N(h)} \sum_{i=1}^{N(h)} [Z(x_i) - Z(x_i + h)]^2 \tag{1}$$

where $N(h)$ is the number of data pairs separated by a given distance $h$; $Z(x_i)$ and $Z(x_i+h)$ denote the ground properties at the locations $x_i$ and $x_i+h$, respectively. Eq. (1) converts the dataset from the first step to discrete data points of variation. In practice, these data points can be fitted using a series of mathematical functions, e.g., spherical, Gaussian, and exponential. The function with the best fitting performance was recognized as the optimal semi-variogram model for the region of interest.

The third step estimates the current ground properties by extrapolating the previously excavated ground properties. The semi-variogram model obtained in the second step can be utilized to determine the weight of extrapolation [28]:

$$\boldsymbol{\lambda} = \boldsymbol{A}^{-1} \cdot \boldsymbol{b} \tag{2}$$

where $\boldsymbol{\lambda}$ is a vector, whose element represents the extrapolation weight of each previously excavated location $\lambda_i$; $\boldsymbol{A}$ is a matrix of the variation in ground properties between any two locations of several previously excavated; and $\boldsymbol{b}$ is a vector of the variation in ground properties between the current location and previously excavated locations. Both $\boldsymbol{A}$ and $\boldsymbol{b}$ can be calculated by the semi-variogram model. With the extrapolation weights from Eq. (2), the current ground properties $Z(x_0)$ can be estimated as a weighted sum of those several previously excavated ground properties $Z(x_i)$:

$$Z(x_0) = \sum_{i=1}^{n} \lambda_i Z(x_i) \tag{3}$$

where $n$ denotes the number of previously excavated locations used for extrapolation. The estimation uncertainty can be quantified via the Kriging variance [29]:

$$Var[Z(x_0)] = 2\sum_{i=1}^{n} \lambda_i \gamma(x_i, x_0) - \sum_{i=1}^{n}\sum_{j=1}^{n} \lambda_i \lambda_j \gamma(x_i, x_j) \tag{4}$$

The Kriging algorithm is a form of Gaussian process regression that models the target variable as a continuous random process [30,31]. Therefore, the prediction from Kriging extrapolation is a random variable that follows a normal distribution, with a mean of $Z(x_0)$ and a variance of $Var[Z(x_0)]$.

**3.2. Utilizing random forest to predict current ground properties**

Random forest algorithm has achieved excellent performance for shield operating parameter analysis [32-35]. Thus, it is adopted here to predict current ground properties from real-time EPB operating parameters. The workflow of RF is described below.

The underlying idea of the RF algorithm is to develop a collection of decision trees and then to average the predictions of all decision trees to provide the prediction. The development of the RF algorithm starts with extracting data subsets from the input of EPB operating data via the bootstrap sampling strategy [36]. This strategy introduces randomness into the sampling process, making the RF algorithm resistant to noise and insensitive to outliers. Then, each data subset is utilized as input to train each decision tree. A decision tree comprises three primary parts: internal nodes, branches, and leaf nodes. The internal node represents a judgment on an EPB operating parameter, the branch indicates the output of a judgment result, and the leaf node denotes the tree prediction for ground properties. The data subset input to a decision tree is classified into leaf nodes in a stepwise manner by executing judgments at each internal node. Each tree in the forest produces an individual prediction of ground properties, and the final prediction of the RF is the average of all tree predictions:

$$F(x_0) = \frac{1}{n} \sum_{i=1}^{n} T_i(x_0) \tag{5}$$

where $T_i(x_0)$ represents the individual prediction of each tree; $n$ represents the number of decision trees in the RF. The uncertainty of RF prediction can be quantified as the variance across all individual tree predictions within the forest [37,38]:

$$Var[F(x_0)] = \frac{\sum_{i=1}^{n}[T_i(x_0) - F(x_0)]^2}{n} \tag{6}$$

### 3.3. Integrating Kriging extrapolation and random forest prediction

The proposed KRF algorithm integrates Kriging extrapolation with RF prediction to obtain a prediction with minimum uncertainty, illustrated in Fig. 2. This is achieved via the weighted least squares method [39]. This method assigns weights to each prediction according to their relative uncertainties. That is, the prediction with a lower uncertainty is given a greater weight, while that with a higher uncertainty is given a smaller weight. The weights of Kriging extrapolation and RF prediction can be calculated as follows:

$$w_{\text{Kriging}} = \frac{Var[F(x_0)]}{Var[Z(x_0)] + Var[F(x_0)]} \tag{7}$$

$$w_{\text{RF}} = \frac{Var[Z(x_0)]}{Var[Z(x_0)] + Var[F(x_0)]} \tag{8}$$

The weighted combination of Kriging extrapolation and RF prediction yields the KRF prediction:

$$\begin{aligned}\hat{Z}(x_0) &= w_{\text{Kriging}} Z(x_0) + w_{\text{RF}} F(x_0) \\ &= \frac{Var[F(x_0)] Z(x_0) + Var[Z(x_0)] F(x_0)}{Var[Z(x_0)] + Var[F(x_0)]}\end{aligned} \tag{9}$$

The pseudocode for the KRF algorithm is summarized in the Algorithm below.

---

**Algorithm: Pseudocode for KRF algorithm**

**Input:** Parameters in Kriging: range ($a$), nugget ($C_0$), and sill ($C+C_0$), Data set R, Forest

1. **Initialization**: model the semi-variogram in Kriging: when $h=0$, $\gamma(h)=C_0$; when $0 < h \leqslant a$, $\gamma(h) = C_0 + C\left(\dfrac{3h}{2a} - \dfrac{h^3}{2a^3}\right)$; when $h > a$, $\gamma(h) = C_0 + C$
2. **for** each set of EPB operating parameters in R **do**
3.     Input operating parameters into *Forest* to predict current ground properties $F(x_0)$
4.     Determine the variance of *Forest*: $Var[F(x_0)] = \left\{\sum_{i=1}^{n}[T_i(x_0) - F(x_0)]^2\right\}/n$
5.     Extract the previously predicted ground properties $Z(x_i)$
6.     Construct the matrix of variation in ground properties **A** and **b** from semi-variogram
7.     Determine the Kriging weights $\lambda_i$ for excavated ground properties: $\boldsymbol{\lambda} = \boldsymbol{A}^{-1} \cdot \boldsymbol{b}$
8.     Extrapolate current ground properties via Kriging: $Z(x_0) = \sum_{i=1}^{n} \lambda_i Z(x_i)$
9.     Derive the Kriging variance: $Var[Z(x_0)] = 2\sum_{i=1}^{n}\lambda_i \gamma(x_i, x_0) - \sum_{i=1}^{n}\sum_{j=1}^{n}\lambda_i \lambda_j \gamma(x_i, x_j)$
10.     Determine the KRF prediction: $\hat{Z}(x_0) = \dfrac{Var[F(x_0)] \cdot Z(x_0) + Var[Z(x_0)] \cdot F(x_0)}{Var[Z(x_0)] + Var[F(x_0)]}$
11. **end for**

**Output:** Current ground properties $\hat{Z}(x_0)$

---

The proposed algorithm integrates the Kriging extrapolations with RF predictions to synergistically predict the current ground properties. The Kriging extrapolation can provide reliable estimates when the ground conditions exhibit a continuous distribution, but it cannot extrapolate new ground properties that have not appeared in the previously excavated ground. By contrast, RF can identify sudden changes in ground conditions from real-time EPB operating parameters, such as faults, but its predictions inevitably include errors because the EPB operating parameters incorporate the subjective actions of machine operators. It is anticipated that these two predictions can synergize under varying ground conditions by adjusting their weights, enabling the proposed algorithm to effectively tackle with both the continuous and sudden changes in ground properties. Note that the proposed algorithm utilizes both real-time EPB operating parameters and ground spatial correlation to predict current ground properties. It is not a time-series model that takes past data points as input to predict the succeeding data points, hence avoiding the time-delayed prediction of EPB operating data, as pointed by Erharter and Marcher [40].

## 4. Implementation of the Proposed Algorithm

### 4.1 Project overview

A tunnel project in Changsha, China, was utilized to assess the reliability and feasibility of the proposed algorithm. The dataset was collected from four tunnel sections of Changsha Metro Line 4; that is, LiuGouLong station to WangYueHu station (LW section), WangYueHu station to YingWanZhen station (WY section), YingWanZhen station to Hunan Normal University station (YH section), and Hunan University station to FuBuHe station (HF section). The lengths of these four sections are 1.57km, 0.8km, 1.23 km, and 1.44km, respectively. The tunnel was excavated using an EPB shield machine with a cutterhead diameter of 6.28 m and an opening ratio of 35%. Precast concrete segments with a width of 1.5 m and thickness of 350 mm constitute the tunnel lining.

The longitudinal geological profile of the tunnel is presented in Fig. 3. In the LW section, the left line traverses strongly, moderately, and slightly weathered slates. In the WY section, both the left and right lines intersect with silty clay, gravel, moderately and slightly weathered slates, and moderately and slightly weathered limestones. In the YH section, the left line primarily passes across completely and strongly weathered mudstone, and moderately weathered limestone, while the right line traverses silty clay, completely weathered mudstone, and moderately weathered limestone. In the HF section, both the left and right lines intersect with strongly and moderately weathered marlite, strongly weathered mudstone, and moderately weathered sandstone. Overall, the geology within the selected region is rather complex and alternating. The tunnel encountered a total of 16 distinct ground types, and the geological parameters of these geomaterials vary in a wide range, summarized in Table 1. The cover depth of the tunnel varies from 6.7 m to 35.5 m. The water table is approximately 5 m below the ground surface.

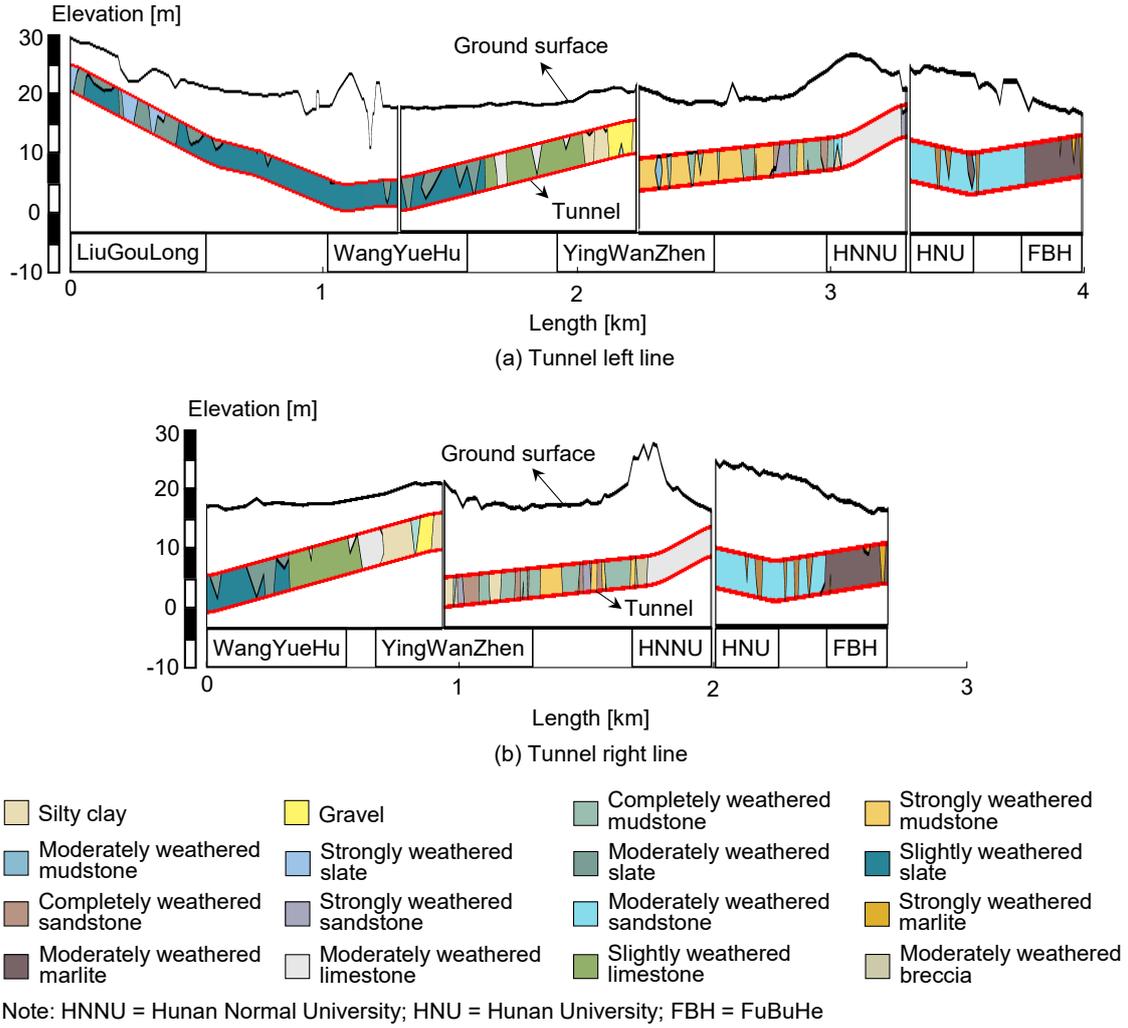

**Fig. 3.** Longitudinal geological profile of the tunnel: (a) Left line, and (b) Right line.

**Table 1.** Geological parameters of the geomaterials in the selected region.

| Geological parameters | Notation | Max.* | Min.* | Ave.* | Unit |
|---|---|---|---|---|---|
| Standard penetration test | $SPT$ | 68.00 | 7.00 | 42.70 | - |
| Dynamic penetration test | $DPT$ | 68.00 | 16.00 | 36.20 | - |
| Uniaxial compressive strength | $UCS$ | 59.39 | 0.06 | 8.64 | MPa |
| Volumetric weight | $\gamma$ | 24.00 | 19.50 | 21.50 | kN/m³ |
| Elastic modulus | $E$ | 4.23 | 1.91 | 2.92 | MPa |
| Poisson's ratio | $\nu$ | 0.34 | 0.22 | 0.25 | - |
| Shear strength | $c$ | 1460.00 | 15.00 | 48.71 | kPa |
| Angle of friction | $\varphi$ | 42.79 | 16.52 | 26.97 | degree |

Note: Max.* is maximum; Min.* denotes minimum; Ave.* is average.

## 4.2 Dataset preparation

The raw dataset consists of the input variable $x$ and the label $y$. The input variable $x$ contains eight EPB operating parameters correlated with ground properties, as elaborated in Section 2. These parameters are automatically captured by the EPB tunneling system at an interval of 1-minute. The statistical description of these parameters is listed in Table 2. The label *y* represents the ground properties currently being excavated. The raw dataset must be preprocessed to enhance the data quality before being input to the KRF algorithm, shown in Fig. 4. The detailed preprocessing procedures are described below.

**Table 2.** Statistical description of input parameters.

| Category | Parameter | Notation | Min.* | Max.* | Ave.* | S.D.* | Unit |
|---|---|---|---|---|---|---|---|
| Hydraulic cylinder | Thrust | $Th$ | 0.28 | 29.79 | 11.67 | 3847.08 | MN |
| | Advance rate | $v$ | 1.00 | 96.38 | 16.45 | 14.75 | Mm/min |
| Cutterhead | Torque | $To$ | 0.50 | 4.73 | 2.16 | 829.15 | MN·m |
| | Cutter rotation speed | $RPM$ | 0.80 | 3.06 | 1.39 | 0.29 | rpm |
| | Penetration | $Pe$ | 1.00 | 75.82 | 12.22 | 11.20 | mm/r |
| Chamber | Chamber pressure | $Cp$ | 0.00 | 3.05 | 1.40 | 0.63 | bar |
| | Foam volume | $Vf$ | 0.00 | 10.35 | 0.53 | 1.39 | L |
| | Water volume | $Vw$ | 0.00 | 154.80 | 17.31 | 43.17 | L |

Note: S.D.* represents standard deviation.

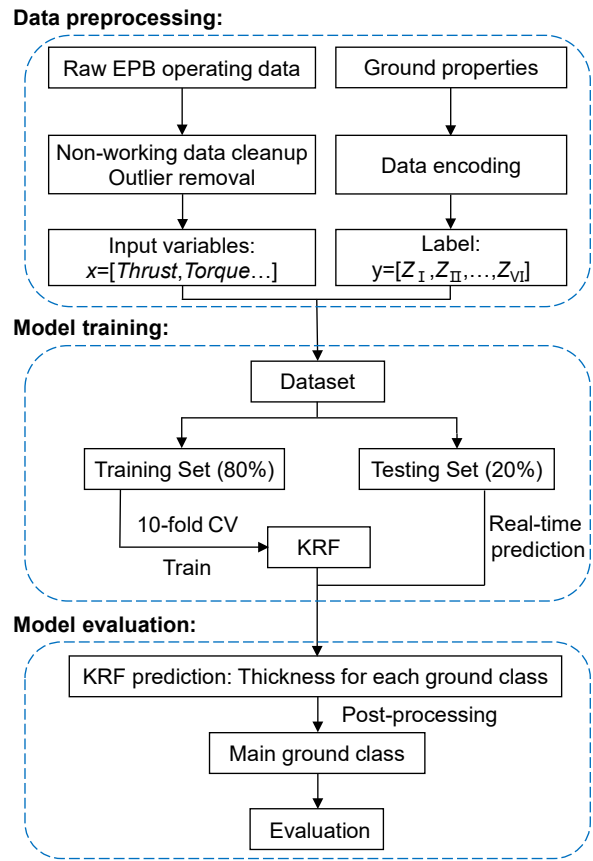

**Fig. 4.** Flowchart for training and evaluating the KRF algorithm.

The preprocessing of input variable *x* removes irrelevant data and noise from the raw EPB operating data, including non-working data cleanup and outlier removal. Non-working data are the data collected when the shield machine stops owing to segmental lining assembly or machine breakdown. These data should be filtered out because they are irrelevant for predicting the ground properties. Specifically, an EPB operating record will be classified as non-working data if any item pertaining to the thrust, torque, advance rate, cutter rotational speed, or penetration is equal to zero [41,42]. The outliers recorded by the EPB tunneling system are typically produced by sensor malfunctions or distortions in data transmission. Here, a box plot method proposed by Tukey [43] was selected to determine the upper and lower limits of normal values and to exclude data outside of the limits.

The preprocessing of label *y* transforms the ground properties into a digitalized format. A data encoding method suggested by Sun et al. [32] was adopted to digitalize the ground properties. Specifically, the encoding method makes a simplification that each stratum is distributed horizontally across the tunnel cross-section. Under the premise of this simplification, the ground properties can be represented by a vector of thickness values for each ground class. Here, the basic quality index (BQ) of the rock mass serves as the indicator for partitioning the ground class [44,45]. It divides soil or rocks into six distinct classes according to the strength and integrity of the rock mass. Table 3 briefly summarizes the definition of the BQ index from both qualitative and quantitative perspectives. Following the preprocessing procedure described above, a dataset containing 205150 samples was established.

**Table 3.** Definition of basic quality index BQ of the rock mass.

| Class | Qualitative characteristics | BQ |
|---|---|---|
| I | Hard rock: complete rock mass | >550 |
| II | Hard rock: relatively complete rock mass<br>Harder rock: complete rock mass | 550-451 |
| III | Hard rock: relatively broken rock mass<br>Harder rock: relatively complete rock mass<br>Softer rock: complete rock mass | 450-351 |
| IV | Hard rock: broken rock mass<br>Harder rock: relatively broken to broken rock mass<br>Softer rock: relatively complete to relatively broken rock mass<br>Soft rock: complete to relatively complete rock mass<br>Compacted or diagenetic cohesive and sandy soils; loess | 350-251 |
| V | Softer rock: broken rock mass<br>Soft rock: relatively broken to broken rock mass<br>All extremely soft rocks and all extremely broken rocks<br>Semi-dry hard and hard plastic clay soil<br>Slightly wet to wet gravel soil and pebble soil | ≤250 |
| VI | Soft plastic clay, saturated silty sand, and soft soil | |

Note: The basic quality index BQ should be determined based on the quantitative measurement index, including the saturated uniaxial compressive strength $R_c$ and rock integrity coefficient $K_v$, as described by the formula $BQ = 100 + 3 R_c + 250 K_v$. The following requirements should be met when using the formula: when $R_c > 90 K_v + 30$, the value of BQ should be calculated by substituting $R_c = 90 K_v + 30$ and $K_v$; when $K_v > 0.04 R_c + 0.4$, the value of BQ should be calculated by substituting $K_v = 0.04 R_c + 0.4$ and $R_c$. (*Standard for Engineering Classification of Rock Masses*, No. GB50128-2014)

The preprocessed dataset was divided into training and test sets for model training and evaluation, respectively, illustrated in Fig. 4. The training set was used to learn the relationship between EPB operating parameters and ground properties. A training set containing diverse ground properties can guarantee the inclusiveness of the trained model. Here, the left line of the LW section, together with the left and right lines of the WY section, as well as the right lines of the YH and HF section, intersect with areas exhibiting more complicated ground properties, demonstrated in Fig. 3. Thus, all the data pertaining to these five lines (80% of the total data) are grouped as the training set. The remaining 20% of the data, i.e., the left lines of the YH and HF sections, are grouped as the test set.

## 4.3 Model training and prediction

The KRF algorithm was trained by calibrating hyperparameters to archive the best performance. The hyperparameters involved in KRF consist of the number of trees, the maximum depth of tree growth (max depth), the minimum number of samples required for internal node splitting (min sample split), the minimum number of samples required for leaf nodes (min sample leaf). These hyperparameters restrict model complexity to prevent overfitting. They were optimized with the aid of a resampling technique called 10-fold cross-validation. In the 10-fold cross-validation, the dataset was partitioned into ten mutually exclusive subsets, followed by ten iterations of training and validation. At each iteration, one was selected for validation, whereas the other nine were used for training. The model performance was assessed by the average error of ten validation subsets. The obtained optimal hyperparameter values are summarized in Table 4, yielding the optimal KRF.

Table 4. Optimal hyperparameters of KRF.

| Hyperparameter | Optional values | Data type | Optimal value |
|---|---|---|---|
| Number of trees | [1~200] | Natural number | 171 |
| Max depth | [1~50] | Natural number | 25 |
| Min sample leaf | [1~50] | Natural number | 10 |
| Min sample split | [2~50] | Natural number | 2 |

The optimal KRF model was validated on the test set to assess its generalization performance. This was achieved by comparing the KRF predictions with labels. To facilitate the comparison, the KRF predictions conducted post-processing to identify the ground class with the largest thickness. That is, the algorithm prioritized the main ground class within the tunnel cross section.

## 5. Results and Discussions

### 5.1. Prediction performance of KRF

The performance of the KRF is assessed by comparing the predicted classes with the actual classes in terms of accuracy and F1-score, confusion matrix, and precision and recall. The accuracy is the proportion of correctly predicted samples relative to the total number of samples and the F1-score is the harmonic mean of precision and recall. These indicators provide a straightforward and intuitive representation of the model performance across the entire dataset. To provide a more detailed assessment of the model prediction performance for each ground class, the confusion matrix, and precision and recall are adopted. The diagonal elements in the confusion matrix represent the number of correctly classified samples, while the off-diagonal elements indicate the number of misclassified samples. A greater number of samples falling on the main diagonal of the confusion matrix suggests a more accurate prediction performance of the model. Precision is the ratio of correctly predicted samples to total predicted samples. Recall is the proportion of actual samples that are predicted correctly. These two indicators reflect the credibility of model predictions and the capacity of the model to detect actual samples, respectively. Higher scores for both precision and recall indicate better overall prediction performance of the model.

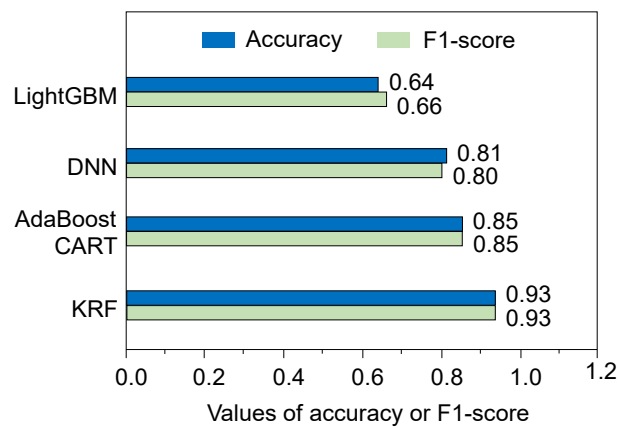

**Fig. 5.** Accuracy and F1-score of four data-driven model predictions on the test set.

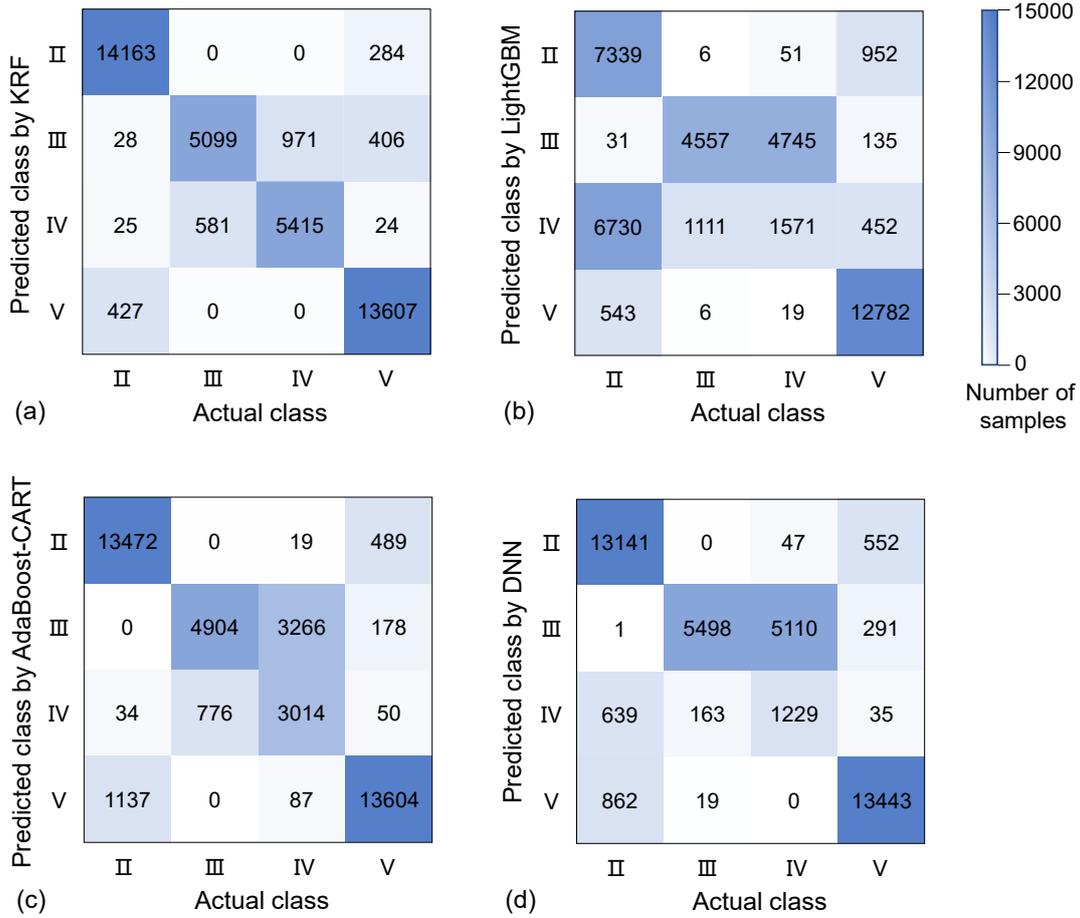

**Fig. 6.** Confusion matrix of model prediction on the test set: (a) KRF, (b) LightGBM, (c) AdaBoost-CART, and (d) DNN.

**Table 5.** Precision and recall values of four data-driven model predictions on the test set.

| Ground class | KRF | | LightGBM | | AdaBoost-CART | | DNN | |
|---|---|---|---|---|---|---|---|---|
| | Precision | Recall | Precision | Recall | Precision | Recall | Precision | Recall |
| II | **0.98** | **0.97** | 0.88 | 0.50 | 0.96 | 0.92 | 0.96 | 0.90 |
| III | **0.78** | 0.90 | 0.48 | 0.80 | 0.59 | 0.86 | 0.50 | **0.97** |
| IV | **0.90** | **0.85** | 0.16 | 0.25 | 0.78 | 0.47 | 0.59 | 0.19 |
| V | **0.97** | **0.95** | 0.96 | 0.89 | 0.92 | 0.95 | 0.94 | 0.94 |

The performance assessment for the KRF algorithm indicates that it can predict the ground properties with an accuracy and F1-score of 93%, illustrated in Fig. 5. Generally, it performs exceptionally in predicting Classes II and V, and moderately well in predicting Classes III and IV, shown in Fig. 6 (a). Specifically, 14163 out of 14643 actual samples in Class II are correctly classified with a precision of 0.98 and recall of 0.97, illustrated in Table 5. Similarly, 13607 out of 14321 actual samples in Class V are correctly categorized, yielding a precision of 0.97 and recall of 0.95. For Classes III and IV, 971 actual samples in Class IV are misclassified as Class III, resulting in a precision of 0.78 for Class III and recall of 0.85 for Class IV.

### 5.2. Comparison among KRF and existing data-driven models

The proposed KRF algorithm is compared with three existing data-driven models, i.e., light gradient boosting machine (LightGBM), classification and regression tree integrated adaptive boosting (AdaBoost-CART), and deep neural network (DNN) [16,46,47], in terms of ground property prediction. These data-driven models are developed on the same dataset to directly predict the main ground class ahead of the EPB shield. They are trained using the 10-fold cross-validation technique, and the optimal hyperparameter selection for each algorithm is summarized in Table 6.

**Table 6.** Optimal hyperparameters of the three existing data-driven models.

| Model | Hyperparameter | Optional values | Optimal value |
| --- | --- | --- | --- |
| LightGBM | Number of estimators | [1~500] | 302 |
|  | Learning rate | [0~1] | 0.45 |
| AdaBoost-CART | Number of estimators | [1~500] | 300 |
|  | Max depth | [1~ 50] | 50 |
|  | Min sample leaf | [1~50] | 4 |
|  | Min sample split | [2~50] | 32 |
|  | Learning rate | [0~1] | 0.1 |
| DNN | Number of hidden layers | - | 2 |
|  | Number of neurons | - | (32, 48) |
|  | Dropout | - | 0.2 |
|  | Learning rate | [0.001~0.1] | 0.001 |
|  | Batch size | [16~128] | 32 |
|  | Epochs | [100~1000] | 1000 |

The prediction performance of the KRF algorithm is compared with the three existing data-driven models via accuracy and F1-score, confusion matrix, precision and recall, receiver operating characteristic (ROC) curve, and precision–recall (PR) curve. A detailed description of the evaluation process is presented below.

The performance comparison of the four data-driven models indicates that the KRF outperforms LightGBM, AdaBoost-CART, and DNN in predicting ground properties, depicted in Fig. 5. The KRF produces the highest accuracy of 93%, surpassing LightGBM, AdaBoost-CART, and DNN by 29%, 8%, and 12%, respectively. Moreover, the F1-score of the KRF also reaches a maximum value of 93%, surpassing the three existing data-driven models by 27%, 8%, and 13%, respectively.

The confusion matrices reveal that the KRF exhibits better performance in distinguishing samples in Classes III and IV than the three existing data-driven models, shown in Fig. 6. Compared to LightGBM, AdaBoost-CART, and DNN, the KRF enhances the precision value for Classes III by 30%, 19%, and 28%, as well as the recall values for Classes IV by 60%, 38% and 66%, respectively. The inaccuracies in distinguishing samples in Classes III and IV for the three existing data-driven models may be attributed to the relatively similar properties of the two ground classes. The proposed KRF algorithm significantly relieves this issue, thus improving the prediction accuracy.

The ROC curves of the KRF and three existing data-driven models are selected to demonstrate the model performance over datasets with various sample distributions, shown in Fig. 7. The ROC curve can reflect the model performance across various datasets because the ROC curve is independent of the proportion of samples from different classes within the dataset [48,49]. A large area under the curve (AUC) implies that the model has a strong capacity to discriminate samples from various ground classes. Apparently, the KRF achieves the best prediction performance among the four data-driven models, with AUC values for all four classes exceeding 0.975. In contrast, the LightGBM exhibited the poorest performance, with AUC values of 0.742, 0.832, and 0.500 for Classes II, III, and IV, respectively. The AdaBoost-CART and DNN perform well in predicting II and V, with the AUC values as high as 0.978 to 0.985. However, the prediction performance of both models for Classes III and IV is subpar. Specifically, the AdaBoost-CART predicts Classes III and IV with AUC values of 0.944 and 0.894, respectively, which are 0.034 and 0.059 lower than those of KRF. The AUC values of DNN in predicting Classes III and IV reach 0.919 and 0.768, which are 0.081 and 0.207 lower than those of KRF. The aforementioned results indicate that the KRF exhibited adaptability to datasets with varying sample proportions.

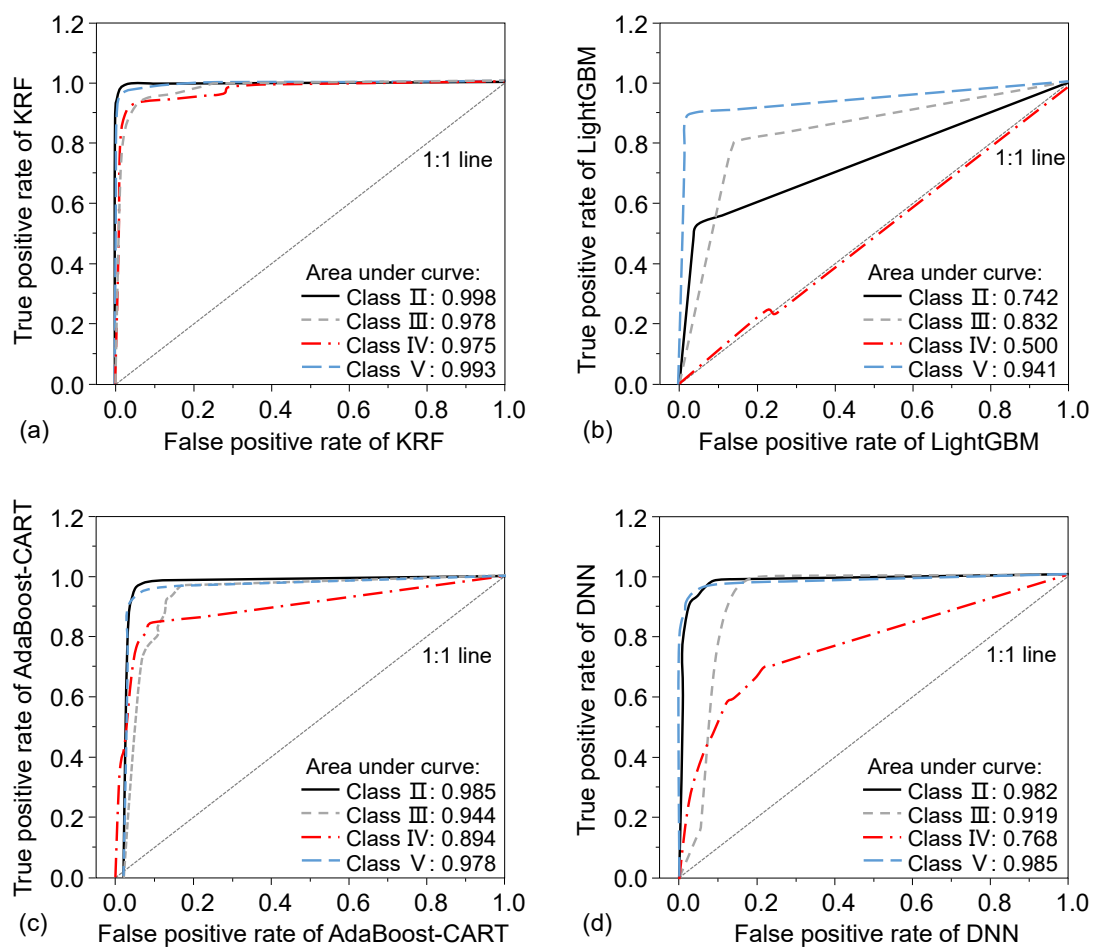

**Fig. 7.** Receiver operating characteristic curves for model predictions on the test set: (a) KRF, (b) LightGBM, (c) AdaBoost-CART, and (d) DNN.

The PR curves of the KRF and three existing data-driven models are illustrated in Fig. 8 to evaluate the model performance in detecting samples in the minority class. The PR curve, which is sensitive to skewness in the dataset, can reflect the model performance on the minority class, thus is used as a supplement to the ROC curve [50,51]. In general, the PR curve of the KRF is higher than that of LightGBM, AdaBoost-CART, and DNN, indicating that the KRF performed better than the three existing data-driven models in identifying the minority class. Specifically, the AUC value of the KRF is 0.935, which is 0.407, 0.115, and 0.218 higher than that of LightGBM, AdaBoost-CART, and DNN, respectively. For engineering applications, samples in the minority class represent abnormal ground conditions, which may trigger construction accidents. The proposed KRF algorithm provides a more effective method for detecting abnormal ground conditions and thus is anticipated to mitigate construction risks.

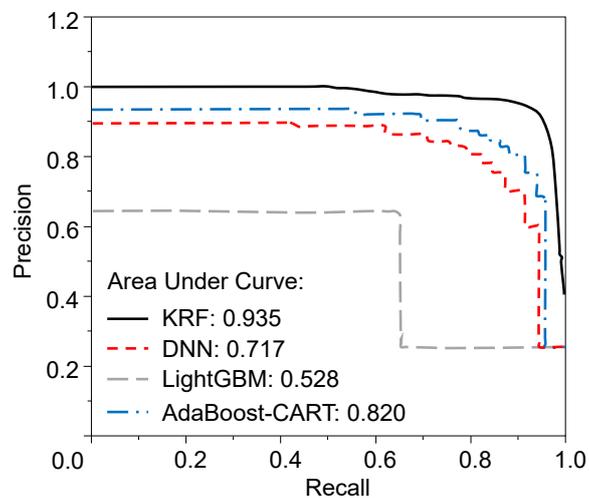

**Fig. 8.** Precision–Recall curves of four data-driven model predictions on the test set.

In conclusion, the proposed KRF algorithm unfailingly outperforms the three existing algorithms of LightGBM, AdaBoost-CART, and DNN in terms of multiple evaluation indicators. The radar chart in Fig. 9 vividly illustrates the performance of KRF, fully enclosing the three existing data-driven models in terms of the six indicators. Specifically, the KRF scores above 0.93 for all six indicators, demonstrating excellent performance across multiple aspects. The AdaBoost-CART ranks second, but its AUC value of the PR curve is the lowest among the six indicators, with a score of 0.82. This indicates that the model bears an insufficient capacity to detect samples in the minority class. Similarly, the DNN scores the lowest on this indicator, ranking third among the four data-driven models. In contrast, the LightGBM performs worst among the four models, with all indicators below 0.75.

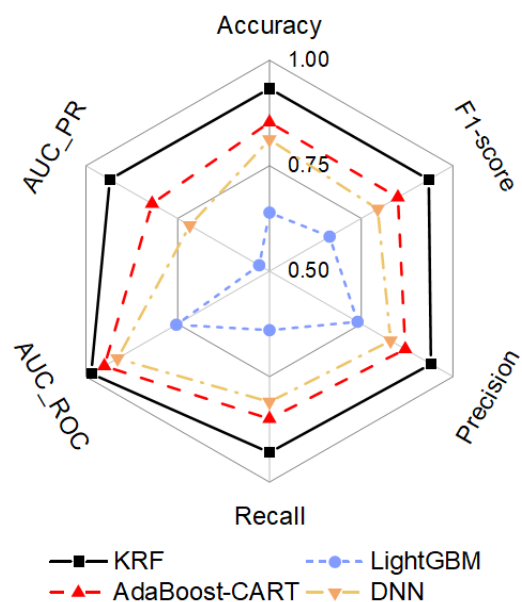

**Fig. 9.** Comparison of prediction performance for four data-driven models on the test set.

### 5.3. Evaluation of transfer performance

One main critique towards data-driven models is that their performance highly relies on the quantity and quality of the dataset. That is, a model trained on a specific dataset may not be readily applicable to another dataset. For the problem of interest in this study, the models are trained on a dataset collected from a project in Changsha city. Therefore, it is important to test if these models can be transferred to datasets from other projects outside of Changsha city. For this purpose, a dataset from the Metro Line 13 project in Shenzhen city is selected to evaluate the four data-driven models' transfer performance. During the evaluation, all the values of the parameters optimized based on the Changsha dataset in the preceding sections will be adopted, and the Shenzhen dataset is directly selected as the test set. Hence, the transfer performance of these models is reflected in their capacity of predicting the new test set from Shenzhen city.

The KRF outperforms the three existing data-driven models in terms of accuracy and F1-score on the Shenzhen dataset, shown in Fig. 10. Specifically, the KRF achieves an accuracy of 89% and an F1-score of 90%, which are 32% and 34%, 20% and 18%, and 24% and 26% higher than those of LightGBM, AdaBoost-CART, and DNN, respectively. On the other hand, the KRF demonstrates robustness in transferring from Changsha dataset to Shenzhen dataset, with only a minor decrease in accuracy and F1-score of 4% and 3%, respectively. In contrast, the three existing data-driven models exhibit significant drops in accuracy and F1-score when transferred from Changsha city to Shenzhen city. The reduction in accuracy and F1-score are 7% and 10% for LightGBM; 16% and 14% for AdaBoost-CART; 16% and 16% for DNN, respectively. Consequently, the proposed KRF algorithm exhibits strong generalization capacities for ground property prediction, exhibiting potential for expansion to other regions.

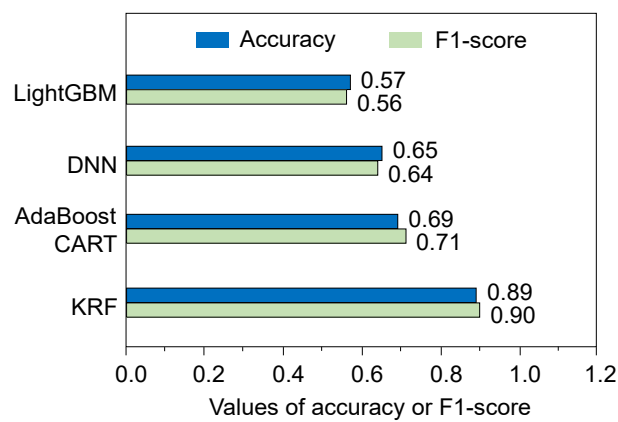

**Fig. 10.** Accuracy and F1-score for predicting Shenzhen dataset with models trained on Changsha dataset.

One of the main sources for the errors of the three existing data-driven models lies in their misclassification of samples with similar properties, particularly those labeled with Classes III and IV. The number of Class III samples correctly classified by LightGBM, AdaBoost-CART, and DNN is only 21, 872, and 107, respectively, shown in Fig. 11. The proposed KRF algorithm can partly overcome this issue, exhibiting better performance in predicting samples in Classes III and IV. Specifically, the KRF correctly predicts 1651, 800, and 1565 more samples for Class III, and 884, 995, and 687 more samples for Class IV, compared to LightGBM, AdaBoost-CART, and DNN.

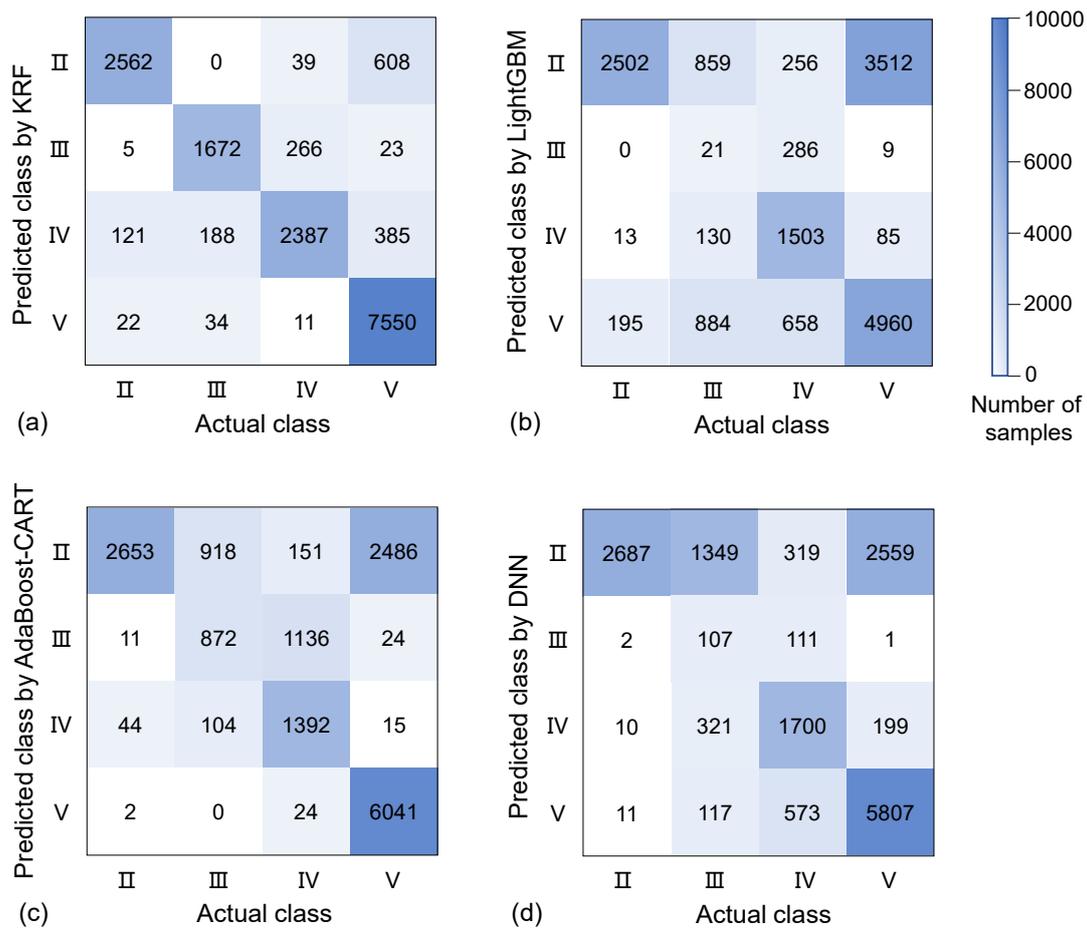

**Fig. 11.** Confusion matrix of predicting Shenzhen dataset with models trained on Changsha dataset for: (a) KRF, (b) LightGBM, (c) AdaBoost-CART, and (d) DNN.

## 5.4. Analysis of parameter importance

A parameter importance analysis is performed to assess the impact of the input parameters on the model predictions. The EPB operating parameters are intertwined manifests of ground properties ahead during EPB tunneling. The interactions among these parameters must be decoupled to recognize the parameter most significant to the ground properties. Here, the importance of each parameter is determined by its contribution to distinguishing samples among distinct ground classes and is normalized to provide a relative score, as proposed by Strobl et al. [52].

The importance analysis of the eight input parameters is shown in Fig. 12. It reveals that the advance rate contributes the most to the prediction of ground properties ahead of the EPB shield. This implies that the alternations in ground properties will be most evident in the differences in the axial resistances ahead. Apart from the advance rate, the chamber pressure is another influential parameter, which works in tandem with the hydraulic cylinder thrust to counterbalance the axial resistances ahead. This observation further confirms the aforementioned perspective. The importance scores of the selected parameters are all greater than 0.03, indicating that the input parameters are appropriately chosen. The ranking of parameter importance reflects the degree of correlation between the EPB operating parameters and ground properties, which can guide shield drivers in prioritizing EPB parameter adjustment.

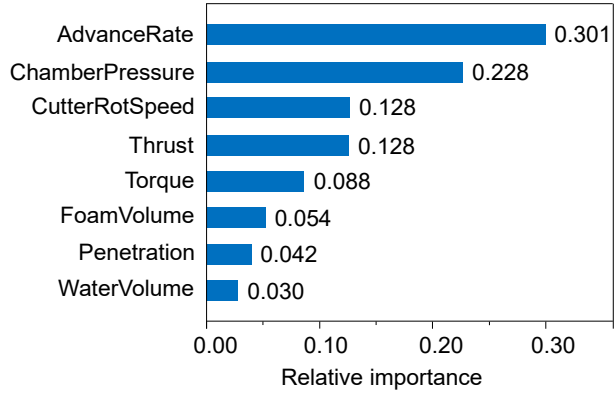

**Fig. 12.** Importance analysis of input parameters.

**Summary and Conclusions**

A kriging-random forest hybrid model is developed for real-time prediction of ground properties ahead of the EPB shield. It incorporates two types of information involved during the EPB tunneling process: prior information and real-time information. Prior information is provided by the Kriging extrapolation of previously predicted ground properties. Its significance to current ground property prediction is embedded in the spatial correlation of ground properties among the current location and previously excavated locations. Real-time information refers to the EPB operating parameters reflecting the shield–ground interaction, which is utilized by random forest to predict the current ground properties. The integration of these two predictions is achieved by assigning weights to each prediction according to their uncertainties, ensuring the prediction of KRF with minimum uncertainty. The applicability of the proposed KRF algorithm is assessed with a dataset from Changsha city. The comparison among the prediction performance of KRF and existing three data-driven models confirms the reliability of KRF. Moreover, the four models trained on the Changsha dataset are further assessed with another dataset from Shenzhen city to evaluate their transfer performance. Major findings are summarized as follows:

1) The proposed KRF algorithm outperforms the three existing data-driven models in terms of ground property prediction. The accuracy of the KRF reaches 93%, which is 29%, 8%, and 12% higher than that of LightGBM, AdaBoost-CART, and DNN, respectively.

2) The proposed KRF algorithm demonstrates a better capacity in classifying the ground classes with similar properties, which is a major challenge for the three existing data-driven models. Compared to LightGBM, AdaBoost-CART, and DNN, the KRF enhances the precision values for Class III prediction by 30%, 19%, and 28%, and for Class IV prediction by 74%, 12%, and 31%, respectively.

3) The proposed KRF algorithm maintains comparable prediction performance when transferring from Changsha city to Shenzhen city. The model reaches an accuracy of 89% and an F1-score of 90%, dropping by only 4% and 3% compared to its performance on the Changsha test set.